\documentclass[12pt]{spieman}

\pdfoutput=1
\usepackage{graphicx}
\usepackage{multirow}
\usepackage{fourier}
\usepackage{indentfirst}
\usepackage{amsmath,amsfonts,amssymb}
\usepackage{setspace}
\usepackage{tocloft}
\usepackage{makecell}
\usepackage{lineno}
\hypersetup{pdfauthor={Zhimin Zhang}, pdftitle={First_Paper}}
\title{Unsupervised monocular stereo matching}
\author[a]{Zhimin Zhang}
\author[a]{Jianzhong Qiao}
\author[a]{Shukuan Lin}
\affil[a]{Northeastern University,  Hunnan Campus, Computer Science and Engineering, No.195, Chong San Road, Hunnan district, Shenyang, China, 110169}

\cftpagenumbersoff{figure}
\cftpagenumbersoff{table}

\begin{document}

\maketitle

\begin{abstract}
 At present, deep learning has been applied more and more in monocular image depth estimation and has shown promising results. The current more ideal method for monocular depth estimation is the supervised learning based on ground truth depth, but this method requires an abundance of expensive ground truth depth as the supervised labels. Therefore, researchers began to work on unsupervised depth estimation methods. Although the accuracy of unsupervised  depth estimation method is still lower than that of supervised method, it is a promising research direction.

 In this paper, Based on the experimental results that the stereo matching models outperforms monocular depth estimation models under the same unsupervised depth estimation model, we proposed an unsupervised monocular vision stereo matching method. In order to achieve the monocular stereo matching , we constructed two unsupervised deep convolution network models, one was to reconstruct the right view from the left view, and the other was to estimate the depth map using the reconstructed right view and the original left view. The two network models are piped together during the test phase. The output results of this method outperforms the current mainstream unsupervised depth estimation method in the challenging KITTI dataset.
\end{abstract}


{\noindent \footnotesize\textbf{*}Fourth author name,  \linkable{zhangzhimin@stumail.neu.edu.cn} }


\section{Introduction}

 With the development of virtual reality and self-driving car etc, depth estimation has  become very hot research. It is also the fundamental problems of the computer vision . At present, the research on depth estimation has achieved very good results. However, most of the research are based on multi-view or binocular of the scene\cite{Furukawa2015Multi} . Such as stereo matching\cite{L'2015Learning} , structure from motion\cite{Sturm1996A} .

 In recent years, with the widespread application of machine learning or deep learning in the field of computer vision, researchers began to apply these methods to the field of depth prediction with a single image \cite{Chen2016Single,Eigen2015Predicting,Ashutosh2009Make3D} or stereo pairs \cite{L'2015Learning,Lecun2016Stereo,Luo2016Efficient} . Although there have been a number of stereo matching articles based on depth learning in recent years, and fruitful results have been achieved both in industry and academia, stereo matching requires costly special-purpose stereo camera rigs. To overcome this problem, researchers began working on depth estimation based on monocular vision. In theory, monocular depth estimation, which does not take into account ground truth depth, is an ill-posed approach for geometric clues, because people who are sensitive to three-dimensional world perception still need two eyes to locate objects in nature. Therefore, it is rather difficult to estimate the depth of the three-dimensional space through a single picture, and the learning model must be a very complicated function.

 At present, supervised monocular depth estimation\cite{Ladick2014Pulling, Li2015Depth,Laina2016Deeper} has undoubtedly become one of the research hotspots of computer vision, and some exciting research results have appeared. The method of these articles is to directly predict the depth value of each pixel of a single image by using deep model, which is  offline trained by the input of single view under the supervision of large number of ground truth depth. Although these studies are fruitful, they need to obtain a large amount of ground truth depths that matching strictly with monocular images through expensive 3D sensors such as LIDAR. Therefore, supervised Monocular depth estimation is a challenging task for collecting vast quantities of corresponding ground truth depth data to training the models. Researchers began to focus on unsupervised depth estimates\cite{Garg2016Unsupervised, Godard2017Unsupervised, Zhou2017Unsupervised} that does not require vast quantities of corresponding ground truth depth data for training. Most of these methods estimate an accurate disparity map by only supervising on the image alignment loss. They rely more on large amounts of high-quality data and effective learning to make deep estimates and require certain geometric constraint. Although this process is difficult to understand and produces suboptimal results, it is also a promising research direction

\begin{figure}
  \begin{center}
    \begin{tabular}{c}
        \includegraphics[height=5.5cm]{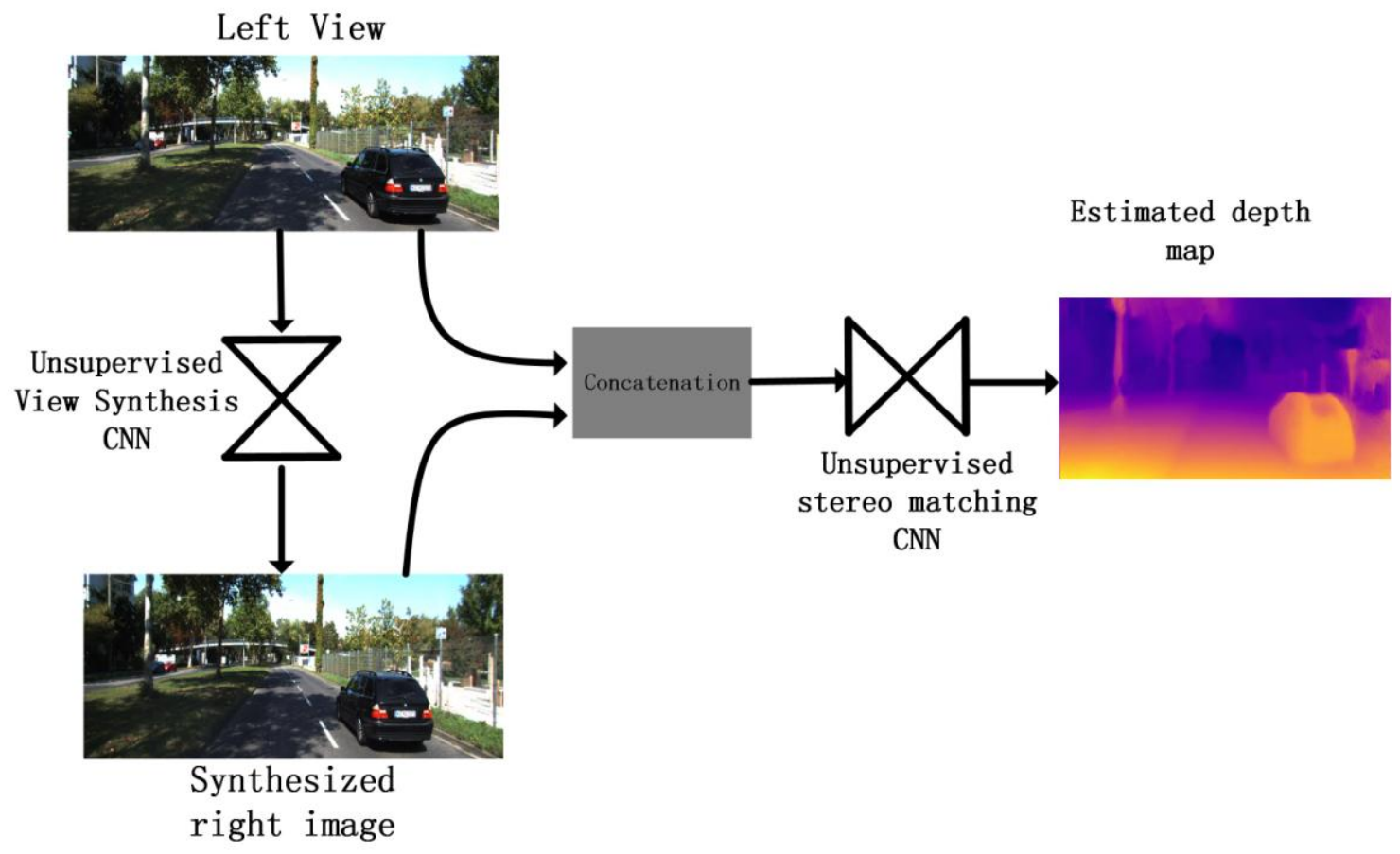}\\
    \end{tabular}
  \end{center}
  \caption{
  \label{fig:DI}Design ideas of our method on monocular stereo matching. We can synthesize right view from single left view by the unsupervised right view synthesis network and then use unsupervised stereo matching network to estimate disparity map from the concatenation input of both left and right views.}
\end{figure}

 Motivated by Luo et al.\cite{Luo2018Single} , our paper proposes an unsupervised stereo matching depth estimation model based on monocular vision. Similarly, we also consider the monocular depth estimation problem as two sub-problems, namely: 1)right view synthesis process; 2) stereo matching process. Unlike Refs.\citenum{Luo2018Single} , which uses semi-supervised, our models purely uses unsupervised depth estimation without ground truth depth and the network architecture that performs end-to-end depth estimation with deep learning network. In order to obtain better depth estimation results, in the training stage, the two models were separately trained, while in the test stage, we connected the two models through pipes and directly estimated the depth value from a single view. A model diagram is shown in Fig.~\ref{fig:DI}.

 In short, we make the following contributions: 1)We proposed an unsupervised depth estimation method from single image. 2) We constructed two deep convolutional network model to achieve our method. Our method is better than the mainstream unsupervised depth evaluation method, and even better than some supervised methods.

\section{Related Work}

 Due to the rise of robotics and virtual reality, depth evaluation has undoubtedly become one of the most popular research points at present. Because machine learning or deep learning has better performance than traditional methods, more and more researchers has applied this methods to depth evaluation and some research results have been achieved. Here we would focus on works related to stereo matching\cite{Luo2016Efficient} and monocular depth evaluation\cite{Ashutosh20083} with machine learning or depth learning, and no assumptions about the scene geometry or types of objects present are made.
\subsection{Stereo Matching}

 The traditional stereo matching algorithm determines the pole by searching the polar geometric line on the stereo pairs. This polar constraint is the basic principle of stereo vision and motion analysis. The binocular views in the stereo matching algorithm are the calibrated images, so the matching problem in 3D space can be transformed into the search problem in 1D space, and obtain the geometrical relation between the depth and the disparity, namely d = fb/z, where the d is the disparity of views, z is the scene depth, f is the camera focal length, the baseline B is the distance of the camera.

 Recently, the vast experiment results show that the stereo matching method based on deep learning outperform using hand defined similarity measures. The methods \cite{Lecun2016Stereo,Pang2017Cascade,Zagoruyko2015Learning} is to learning the matching function through the process of finding pixels points consistent with the left view from the right view of stereo pairs. Luo et al. \cite{Luo2016Efficient} proposed a faster and more accurate depth estimation network architecture. The architecture consists of two Siamese network and product layer that computes the inner product of feature vectors from two Siamese network. This method is treating disparity estimation as a multi-classification problem, that is ,every possible disparity as a class. L'ubor et al.\cite{L'2015Learning} proposed stereo matching architecture based on convolutional neural network with the ground truth disparity to constructing a binary classification dataset. The approach focuses on the matching cost computation by learning a similarity measure on small image patches. Mayer et al.\cite{Mayer2016A} presented an novel deep CNN network with fully convolutional\cite{Long2015Fully} to achieve end-to-end training process using synthetic stereo pairs, called DispNet. The network architecture of FlowNet\cite{Dosovitskiy2015FlowNet} is similar to DispNet\cite{Mayer2016A} , which are also applied to optical flow estimation. Pang et al.\cite{Pang2017Cascade} proposed a cascade residual convolutional neural network architecture composing of two stages. The two stages, which can generate residual signals across multiple scales, include improved DispNet\cite{Mayer2016A} by add additional up-convolution modules, and the network of explicitly rectifying the disparity. Although the above methods based on learning outperformed traditional stereo matching methods, they rely on vast accurate ground truth disparity data and stereo image pairs at training time.
\subsection{Monocular Depth Estimation}

 The stereo matching method has certain requirements for binocular camera, which is not suitable for the actual single camera equipment. Therefore, researchers began to study the monocular depth estimation and has obtained a series of research results. For supervised learning depth estimation, Saxena et al.\cite{Ashutosh20083} proposed first supervised learning approach to resolve the problem of depth estimation from monocular images. The model adopted a discriminatively-trained MRF with multi-scale local and global image features, and models the depth of each point and the depth relation of different points. With the widely application of CNN in computer vision, researchers began to apply the deep learning method to monocular depth estimation. Eigen et al.\cite{Eigen2014Depth} was the first article that attempted to solve the monocular depth estimation problem using CNN architecture by employing two network models that one network model makes coarse global prediction for entire image and another network models refines this prediction locally. The loss function of this model is adapted a scale-invariant error. Subsequently, the authors improved the network and generated a new multi-scale CNN network architecture\cite{Eigen2015Predicting} with fully convolutional up-sampling network\cite{Long2015Fully} , which can complete three visual tasks, including depth prediction, surface normal estimation, and semantic labeling. Laina et al.\cite{Laina2016Deeper} proposed a fully convolutional residual network\cite{Kaiming2015Deep} to model the mapping relation between a single images and ground truth depths. The architecture adapted a novel up-sampling model called up-projected to improve the output resolution and introduced the reverse Huber loss to optimize the network. Liu et al.\cite{Liu2016Learning} proposed a deep learning model based on deep CNN and continuous CRF for estimating monocular depth. On the basis of this, the author further proposes an equally effective model based on FCN and a new superpixel pooling method to speedup the patch-wise convolutions in the estimation model.

 Although supervised learning can achieve well results of depth estimation, this learning method requires vast ground truth depth data, which are difficult to obtain for practical application. To overcome this problem, researchers began to focus on unsupervised depth estimation. Xie et al.\cite{Xie2016Deep3D} proposed a unsupervised transformation method of 2D to 3D of films, which is essentially a method of reconstructing the right view based on single left view by extracting stereo pairs from existing 3D films as supervision training. This model predicted a probabilistic disparity-like map and combined it with left view to reconstructed right view. Garg et al.\cite{Garg2016Unsupervised} proposed a unsupervised deep model based on polar geometry to implement end-to-end monocular depth estimation by only supervising on the image alignment loss. However, the author adapted Taylor expansion to linearize the not fully differentiable loss function. Godard et al.\cite{Godard2017Unsupervised} proposed a novel unsupervised depth estimation model based on proposed the model by \cite{Garg2016Unsupervised} and adapted a new fully differentiable appearance matching loss and left-right disparity consistency loss. Due to the sparsity of ground truth depth data acquired by radar, the supervised learning cannot accurately estimate the image depth, so Kuznietsov et al.\cite{Kuznietsov2017Semi} proposed semi-supervised depth estimation model, which can make unsupervised learning on dense correspondence field and use sparse radar depth data for further supervised learning. For explicitly imposing geometrical constraint, Luo et al.\cite{Luo2018Single} decomposed the monocular depth estimation into two sub-problems for the first time that one is view synthesis procedure and another is stereo matching. Similar to the semi-supervised method, the stereo matching network also needs sparse ground truth disparity data for the supervised learning.

\begin{figure}
\begin{center}
    \begin{tabular}{c}
        \includegraphics[height=5.5cm]{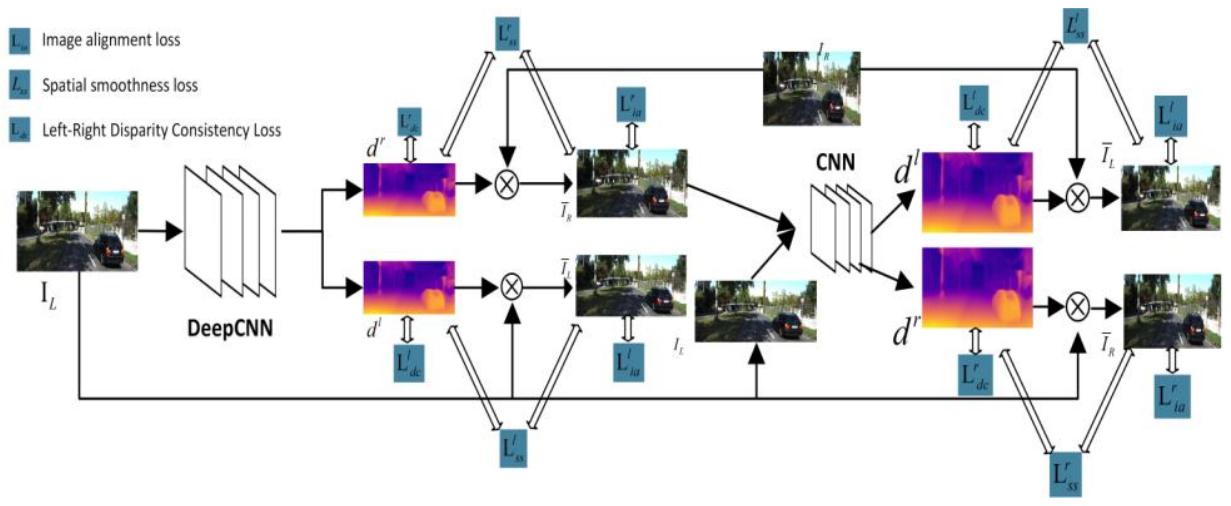}\\
    \end{tabular}
\end{center}
  \caption{The implementation procedure of our unsupervised monocular stereo matching .The network model consists of two parts, namely right view synthesis network and stereo matching network. The input original left view is first processed by CNN based on Resnet50 and FCN of upsampling to reconstruct right view .Then the concatenation of both synthetic right view and original left view input the stereo matching network composed by vgg16 and upsampling FCN to estimate an accurate disparity.}\label{fig:IP}
\end{figure}

 We can see the comparison results from Table~\ref{tab:comp_mon_bin}, that the stereo matching models outperforms monocular depth estimation models under the same unsupervised depth estimation model. So inspired by Luo et al.\cite{Luo2018Single} , we proposed a unsupervised monocular image stereo matching model that composed by the view synthesis network and stereo matching network. For these two network, we was suggested from Refs.\citenum{Godard2017Unsupervised} , that constructed an unsupervised end-to-end convolutional network model with similar structure. We can synthesize right view from left view through view synthesis network that was trained by the loss of consistency between the predicted view and the original image. Then we input the concatenation of both original left and synthesized right views into stereo matching networks for depth estimation . The implementation procedure of our unsupervised monocular vision stereo matching is illustrated in Fig.~\ref{fig:IP}.

\begin{table}[ht]\normalsize
\setlength{\tabcolsep}{0.5mm}{
  \centering
    \caption{Comparison of effects between monocular and binocular inputs from the same unsupervised network model. Where K is the KITTI dataset. The experimental results show that the stereo matching models outperforms monocular depth estimation models under the same unsupervised depth estimation model. }\label{tab:comp_mon_bin}
    \begin{tabular}{|c|c|c|c|c|c|c|c|c|c|}
    \hline
    \multirow{2}*{Method}&\multirow{2}*{Dataset}&\multirow{2}*{Type}&RMSE&RMSE(log)&ARD&SRD&$\delta<1.25$&$\delta<1.25^2$&$\delta<1.25^3$\\
    \cline{4-10}
    &&&\multicolumn{4}{|c|}{--lower is better--}&\multicolumn{3}{|c|}{--higher is better--}\\
    \hline
    ours with VGG16&\multirow{4}*{K}&Mono&6.125&0.217&0.1235&1.3882&0.841&0.936&0.975\\
    \cline{1-1}\cline{3-10}
    ours with ResNet50& &Mono&5.764&0.203&0.114&1.246&0.854&0.947&0.979\\
    \cline{1-1} \cline{3-10}
    ours with VGG16& &Bino&4.434&0.146&0.0669&0.899&0.947&0.978&0.988\\
    \cline{1-1}\cline{3-10}
    ours with ResNet50& &Bino&4.593&0.150&0.0701&1.0391&0.946&0.977&0.988\\
    \hline
    \end{tabular}}
\end{table}

\section{Approach}

 This section describes our unsupervised monocular stereo matching model. We describe the model in two parts that one is the right view synthesis network and the other is the stereo matching network. In the training stage, in order to get better output results for each model, we trained the two models separately, but stereo matching network need synthetic right view as the training data. In the testing stage, we loaded two models successively and the output results of the former network are transferred to the latter network as input data through the pipeline.

\subsection{Unsupervised Monocular Depth Estimation}

 Depth map is the actual distance between each pixel in a 2D image and the camera that took the image. The so-called monocular depth estimation is that given a 2D image$I$, we use a function $f$ to predict the depth $z$ corresponding to each pixel in the image. The process can be described as:$z=f(I)$. The current monocular depth estimation method based on supervised learning uses the single RGB image as the input, and the ground truth depth data as labels training the neural network to construct the fitting function $f$, so that the scene depth information can be obtained according to the single image. However, this method needs to know the expensive ground truth depth data corresponding to the input images as the reference for training.
 For unsupervised monocular depth estimation, it is generally to estimate the disparity map from the input left view through the depth CNN network, and then the generated disparity map and the original view are used to reconstruct the left and right views. In the training stage, we obtain the depth estimation model by optimizing the original view and the image alignment loss function of the reconstructed view or other additional loss functions. In the test stage, the image depth can be estimated by directly inputting a single view based on the trained model. Our method uses the same design idea with other unsupervised model. But the experimental results show that the stereo matching models outperform monocular depth estimation models under the same unsupervised depth estimation model. So we transformed the depth estimation problem into the process of image synthesis and stereo matching based on unsupervised learning. We use the image synthesis network to learning the function that it can reconstructed the right view from left view. Then the unsupervised stereo matching network can train the convolutional network to estimate disparity from the input of combination of left and right views. The stereo matching model refers to that the input data of this model is the concatenation of both left views and right views. In order to verify which training method can get better results for stereo matching model, we adapted three ways to train that it can be see from experimental section.

 As we can see from Fig.~\ref{fig:IP}, at training time of view synthesis network, we have access to a lot of pairs of calibrated stereo pairs$I^l$ and $I^r$, with the left view $I^l$ as the input data and the stereo pairs $I^l and I^r$ as the supervision labels. The input left view $I^l$ is processed by convolutional neural network to find the dense correspondence field $d^r$ that was the disparity map of the right view relative to left view. Then, we can reconstructed right view $\bar{I}^r$ by the bilinear sampling function $I^l(d^r)$. We can training the loss function by supervising on the image alignment loss between the original right view $I^r$ and synthesis right view $\bar{I}^r$ to generate view synthesis network model. Similarly, we can also use the convolutional neural network to process input left view $I^l$ for obtaining the dense correspondence field $d^l$ that was the disparity map of the left view relative to right view. Then,we can synthesize the left view given the right view by the bilinear sampling formula $\bar{I}^l=I^r(d^l)$ and optimize the alignment loss function between the original view and the synthetic view. During testing, we just need to input the left view into the trained network model to estimate the right view, instead of stereo pairs like the input in training stage. Then we can input the concatenation of both left and synthetic right views into the stereo matching network to estimate depth with similar principles as described above.

\subsection{Network Architecture}

 In order to select a more suitable encoding architecture, we have done relevant experiments on view synthesis network and stereo matching network with vgg16 and ResNet50 networks respectively as Table ~\ref{tab:comp_mon_bin}.

 The experimental results show that the evaluation metrics of ResNet50\cite{Kaiming2015Deep} outperform that of VGG16\cite{simonyan2014very} network in view synthesis process. Therefore, we use ResNet50 network as the encoder part of the view synthesis network, and The full convolution network is used to replace the full connection layer as the decoding part of the network. View synthesis network architecture is shown in Table ~\ref{tab:Syn_net}. Our input image is the RGB left view with 256*512 resolution size. In the coding part, 2048 feature images with a resolution of 4*8 are extracted after 50 convolution operations. In order to obtain accurate depth information, we adapted skip connections to concatenate both different scales feature map of encoder parts and same resolution feature maps of decoder parts for up-sampling, which can get disparity map by simple computation. Finally, the right view is synthesized based on the polar geometry by estimating the depth information.

\begin{figure}
\begin{center}
    \begin{tabular}{c}
        \includegraphics[height=3.5cm]{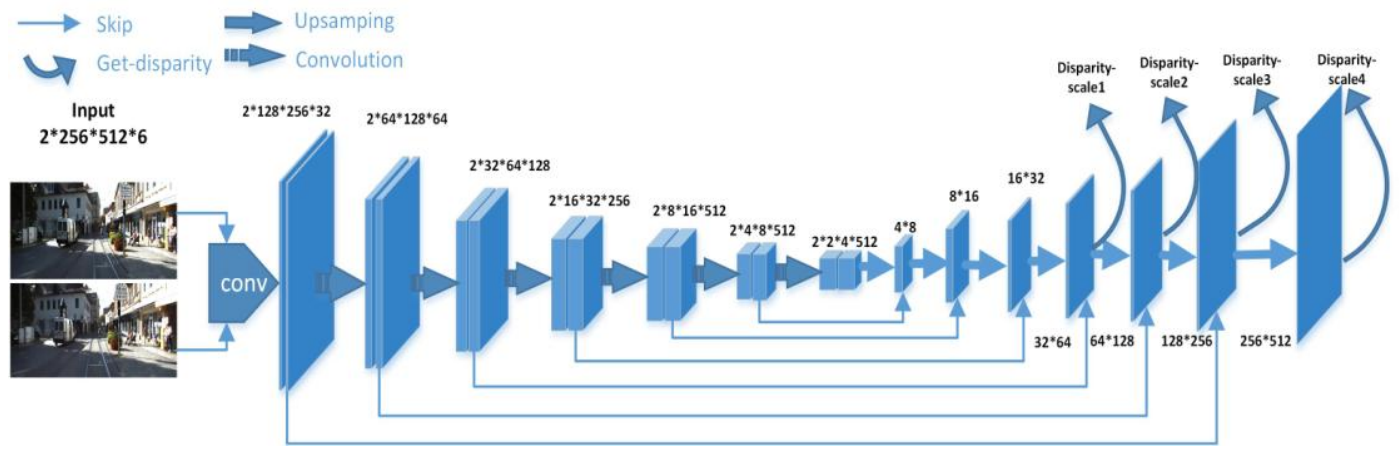}\\
    \end{tabular}
\end{center}
  \caption{Stereo matching network encoder-decoder architecture. The input of model was the concatenation both original left views and synthesized right views. We input the output of encoder network at each resolution to decoder network at the same scales layers by skip connections. We selected four scales high resolution up-sampling outputs to calculate disparity map.}\label{fig:st_arch}
\end{figure}

 As can be seen from Table ~\ref{tab:comp_mon_bin}, for the stereo matching network, VGG16 as the encoding part is better than the encoding result of ResNet50. Fig.~\ref{fig:st_arch} is the network architecture diagram of our stereo matching network. The input is the concatenation of both original left views and synthesized right views with 256*512 resolution size. After the coding of VGG16 network, 512 feature maps with a resolution of 2*4 were finally obtained. In the same way, the feature map of seven scales is up-sampled by the method of skip connection. Then the disparity map is calculated by selecting the feature graph of four high resolution scales as the model optimization factor.



\begin{table}[ht]\footnotesize
  \caption{View synthesize network encoder-decoder architecture. The res\_convx\_3 refers to the convolution block of our deep residual network, which include three convolution process each block. The upconv\_block was the upsampling convolution block which the input was the concatenation both different scales feature map of encoder parts and same resolution feature maps of decoder parts.      }\label{tab:Syn_net}
\setlength{\tabcolsep}{0.5mm}{
  \begin{center}
    \begin{tabular}{c c c}
    \hline
    Layer&Output(resolution*channels)&Inputs\\
    \hline
    $conv^7_2$&128*256*64&Input(RGB)\\
    $max\_pool^3_2$&64*128*64&conv\\
    \hline
    res\_conv1\_3&64*128*64&$max\_pool^3_2$\\
    res\_conv2\_3&64*128*64&res\_conv1\_3\\
    res\_conv3\_3&32*64*256&res\_conv2\_3\\
    \hline
    res\_conv4\_3&32*64*128&res\_conv3\_3\\
    res\_conv5\_3&32*64*128&res\_conv4\_3\\
    res\_conv6\_3&32*64*128&res\_conv5\_3\\
    res\_conv7\_3&16*32*512&res\_conv6\_3\\
    \hline
    res\_conv8\_3&16*32*256&res\_conv7\_3\\
    res\_conv9\_3&16*32*256&res\_conv8\_3\\
    res\_conv10\_3&16*32*256&res\_conv9\_3\\
    res\_conv11\_3&16*32*256&res\_conv10\_3\\
    res\_conv12\_3&16*32*256&res\_conv11\_3\\
    res\_conv13\_3&8*16*1024&res\_conv12\_3\\
    \hline
    res\_conv14\_3&8*16*512&res\_conv13\_3\\
    res\_conv15\_3&8*16*512*&res\_conv14\_3\\
    res\_conv16\_3&4*8*2048*&res\_conv15\_3\\
    \hline
    upconv\_block1&8*16*512&res\_conv16\_3\\
    \hline
    upconv\_block2&16*32*256&upconv\_block1\\
    &&res\_conv13\_3\\
    \hline
    upconv\_block3&32*64*128&upconv\_block2\\
    &&res\_conv7\_3\\
    \hline
    upconv\_block4&64*128*64&upconv\_block3\\
    &&res\_conv3\_3\\
    \hline
    upconv\_block5&128*256*32&upconv\_block4\\
    &&$max\_pool^3_2$\\
    \hline
    upconv\_block6&256*512*16&upconv\_block5\\
    &&$conv^7_2$\\
    \hline
    \end{tabular}
    \end{center}}
\end{table}

\subsection{Loss Function}

 We adopted the same unsupervised optimization method for view synthesis network and stereo matching network as shown in Fig.~\ref{fig:IP}, that is, we acquire the disparity map by learning the deep network, and reconstruct the left and right views according to the geometric relationship between the disparity map and the stereo pairs. Therefore, we use the same loss function for the two network models.
\subsubsection{Related Formula}
 In order to obtain the formula of loss function, we explain the derivation process of the related formula based on the previous theory. We use the rectified stereo pairs as training data. Given the camera's internal and external parameters, we assume photo-consistency between the left view and the right view. According to the geometric properties and polar constraints of the binocular camera, the corresponding formula can be obtained.

\begin{equation}\label{1}
    \begin{split}
        &\bar{I}^l(i,j)=I^r(i,j-d^l_{i,j})\\
        &\bar{I}^r(i,j)=I^l(i,j-d^r_{i,j})
    \end{split}
\end{equation}

 Where$(i,j)\in \Omega$, $\Omega$ is the image space of $I$, and i,j refer to the horizontal and vertical coordinates of the pixel position of the image. We can also get the depth estimation $Z=\frac{bf}{d}$ by given the baseline distance b between the left and right cameras, the cameras focal length f and the disparity map.

\subsubsection{Loss Function of Model }

 Our network model adopts the unsupervised network model, which takes the original left image or the concatenation of the original left image and the composite right image as inputs, and there is no ground truth depth data as the supervised label in the training stage. Therefore, we are inspired by Refs~\citenum{Godard2017Unsupervised} to estimates the disparity map by optimizing image alignment loss, left-right disparity consistency loss and spatial smoothness loss. We define a loss $L_{\theta}$ as the total loss of the different constraint forms for two network model.

\begin{equation}\label{2}
  L_{\theta}=\alpha L_{ia} + \beta L_{ss} + \gamma L_{dc}
\end{equation}

 Where $L_{al}$ was the image alignment loss of reconstructed view and original view, $L_{sm}$ was the regularization term on the spatial smoothness of disparity values. Each loss terms contains two loss formulas that was the respective image alignment loss between the reconstructed left and right views and the original views. $\alpha,\beta,\gamma$ are used as the adjustment parameter of specific gravity of each loss function.

\textbf{Image alignment loss:}
 From article Refs.\citenum{Zhao2015Is} , we can see that the optimization results of the combined features of MS\_SSIM \cite{Zhou2004Image} and L1 loss functions were better than the single function of these two loss functions. In order to optimize the quality of the reconstructed views, we use the mixed loss function came from L1 and MS\_SSMI as our photometric image reconstruction cost function $L_{ia}$, which calculates the alignment loss between original stereo pairs $I^l,I^r$ and reconstructed stereo pairs $\bar{I}^l,\bar{I}^r$.

\begin{equation}\label{3}
    \begin{split}
        L_{ia}=\frac{1}{N}\sum_{n\in{(l,r)}}\sum_{i,j}\gamma L^{MS\_SSIM}(I_{i,j}^n,\bar{I}_{i,j}^n)+ (1-\gamma)L^{l1}G_{\sigma_G^M}(I_{i,j}^n,\bar{I}_{i,j}^n)
    \end{split}
\end{equation}

 where \\
 \begin{equation}\label{4}
    \begin{split}
       &L^{MS\_SSIM}(I_{i,j}^n,\bar{I}_{i,j}^n)=1-MS\_SSIM(I_{i,j}^n,\bar{I}_{i,j}^n),\\
       &L^{l1}(I_{i,j}^n,\bar{I}_{i,j}^n)=|I_{i,j}^n-\bar{I}_{i,j}^n|\\
    \end{split}
 \end{equation}
Here,we selected Gaussian smoothing kernel$G_{\sigma_G^M}$ with normal distribution of $\sigma=1px$ for $L^{l1}$ and set the parameter $\gamma$ to 1

\textbf{Spatial smoothness loss:}
 It is well known that the disparity estimation problem is ill-posed in homogeneous regions of the scene without ground truth disparity and depth discontinuities often occur at image gradients. Thus as suggested in this paper\cite{Kuznietsov2017Semi} , we add the edge edge preserving regularizer as part of the loss function using the image gradients $\partial I$,with the $n\in{l,r}$.

\begin{equation}\label{5}
  L_{sm}=\frac{1}{N} \sum_n \sum_{i,j}|\partial_x d_{ij}^n|e^{-|\partial_xI_{ij}^n|}+|\partial_y d_{ij}^n|e^{-|\partial_yI_{ij}^n|}
\end{equation}

\textbf{Left-Right disparity consistency loss:}
 We refer to the loss of the article \cite{Godard2017Unsupervised} , which is based on the geometric constraint of stereo to able to make the left-right disparity map convert to alignment loss. To ensure consistency, we adopted a simple L1 loss as left-right disparity consistency loss $L_{ds}$. By optimizing the loss function, we can get more accurate disparity maps.

\begin{equation}\label{6}
  L_{ds}=\frac{1}{N} (\sum_{i,j}|d_{i,j}^l-d_{i,j+d_{i,j}^l}^r|+\sum_{i,j}|d_{i,j}^r-d_{i,j+d_{i,j}^r}^l|)
\end{equation}

\section{Experiments}

 Currently, the commonly used data sets reconstructed in 3D scenes include indoor data NYU Depth Dataset \cite{Silberman:ECCV12} , outdoor Dataset Make3D \cite{Ashutosh2009Make3D} and KITTI Dataset \cite{Geiger2012Are} in self-driving scenes. Our research objective is to study the automatic driving of cars under complex road conditions. Therefore, this section shows our experiments and results that we compare the performance between our approach and current state-of-the-art monocular depth estimation method on the popular KITTI dataset. In this section, we also discover and prove the theoretical correctness of our method.

\subsection{Dataset}

 Kitti dataset is the most widely used image dataset in the field of autonomous driving. The dataset\cite{Geiger2013Vision} records the six-hour traffic scenarios by a series of sensors, including high-resolution color and gray stereo cameras, a 3D laser scanner, and a high-precision GPS/IMU inertial navigation. The scenarios are captured by driving around inner city of Karlsruhe city on high-speed, in rural areas, with many static and dynamic objects. This dataset is calibrated, synchronized and timestamped, and we provide the rectified and raw image sequences.

 We evaluate our method with rectified stereo pairs from 61 scenarios of the KITTI dataset which include the categories "city","residential" and "road". In order to better show the comparison results of our method and other methods, our experiment referred to the data allocation scheme proposed by Eigen et al.\cite{Eigen2014Depth} . We randomly selected 28 scenes from all 61 scenarios, and then randomly selected 697 images from them as test data. The remaining 33 scenes contained a total of 30159 images, of which 29,000 were used for training data and the rest for verification data.

\subsection{Implementation Details}

 To better training our model, we implement our model in Tensorflow \cite{Abadi2016TensorFlow} on the experimental platform of 32GB E5-2620v4 with 12GB NVIDIA GTX 1080Ti. Since the input of the stereo matching network needs to have a high-quality input view from the concatenate of original left view and synthesized right view, in order to get a better reconstructed right view, we trained the two networks separately.

\textbf{View synthesis network:}
 For the training of view synthesis network, we adopt ResNet50 \cite{Kaiming2015Deep} as encoder network and use fully convolutional network with bilinear sampler as decoder network. we initialize the weights of the encoder network using the trained ResNet50 model from ImageNet and other weights using random initialization from the gaussian distribution with a standard deviation of 0.01. The network model contains 48 million trainable parameters with the input resolution 512x256. We set the default batch size to 10 and the default epochs number to 60. In order to make the model converge quickly, we set the initial learning rate to 0.0001, which for the first 40 epochs, we kept the constant learning rate, and then reduce it by a factor of 2 after every 10 epochs until the end. During optimization, we refer to the paper\cite{Godard2017Unsupervised} that we set the parameters $\alpha = 1$ ,$\beta = 1$ and $\gamma=0.1 $ of the loss function and constrain the output disparity to be between 0 and $d_{max}$, where $d_{max}$ is assigned 0.3 multiply by the resolution width of output image. In order to estimate the disparity of images at different scales, we obtain four different scales of images by down-sampling of a factor of two.

\textbf{Stereo matching network:}
 Our stereo matching network architecture is similar to the view synthesis network architecture and the input data is the concatenation of both left view and synthesis right view, where the synthesis right view has the same resolution as the original left view by the view synthesis network's up-sampling. Therefore, we refer to the basic setup of the view synthesis network to set stereo matching network. However, due to certain errors in the reconstructed right view, in order to avoid large shocks and overfitting, we reduced the learning rate to 0.00001 and increased the number of epochs to 80. We find that for stereo matching algorithm, VGG16\cite{Karen2014Very} network model is better than ResNet50 through experimental verification from Table~\ref{tab:comp_mon_bin }, so we adopt the VGG16 network model as the encoding part, and the decoding part is still the full convolution network with bilinear interpolation sampling. We initialize the weights of the encoder network using the trained VGG16 model from ImageNet and other weights using random initialization from the gaussian distribution with a standard deviation of 0.01. Although we use the synthesized right view as part of the input data of the stereo matching model, the supervised image of the image alignment loss still use the original stereo pairs, and more accurate depth estimates can be obtained.

 We abandoned the addition of batch regularization in the two network models because experiments showed that the structure did not play an important role in the experimental results. We augment the image data during data loading. We flip and swap every stereo pairs in equal probability, and make sure the both images are in the right position relative to each other. At the same time, we also adjust the brightness, contrast and color of the stereo pairs by making linear changes to the pixel from uniform distribution in the range [0.8,1.2] for each color channel, [0.8,1.2] for gamma, [0.5,2.0] for brightness.

\begin{figure}
\begin{center}
    \begin{tabular}{c}
        \includegraphics[height=3.5cm]{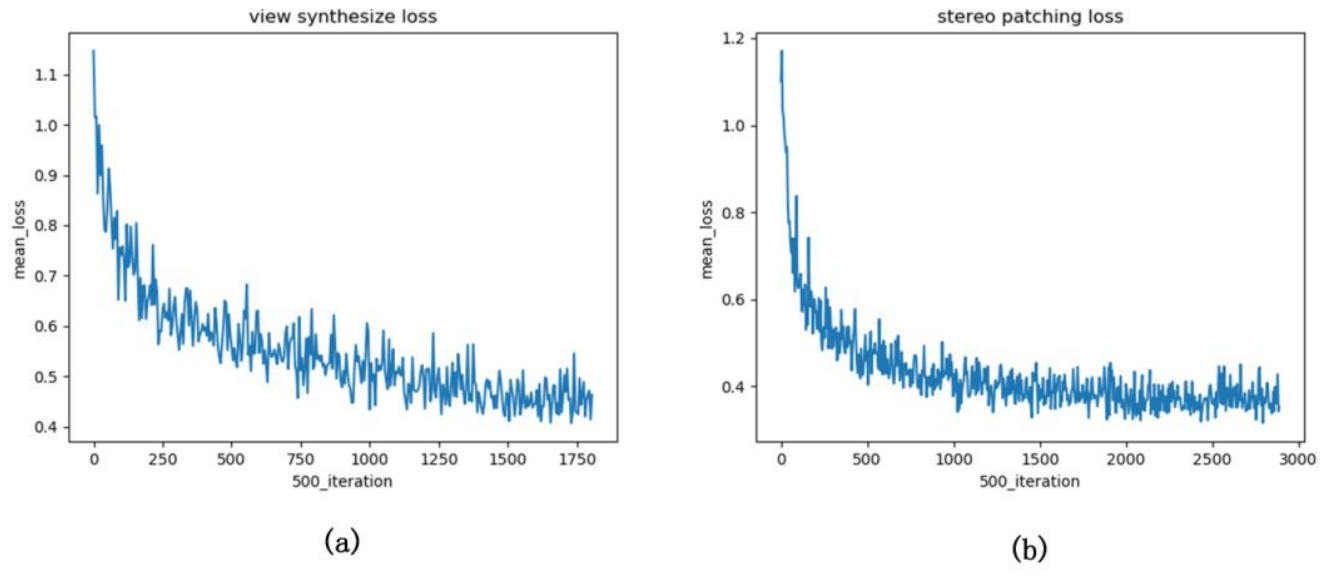}\\
    \end{tabular}
\end{center}
  \caption{The optimization process of the loss function of our view synthesis network(a) and stereo matching network(b) in training}
  \label{fig:loss}
\end{figure}

 \textbf{Training loss:}
 Fig.~\ref{fig:loss} shows the optimization process of the loss function of our view synthesis network and stereo matching network in training. In order to better show the training trend of the loss function as a whole, we select a loss value for every 100 iterations and make an average value for every five iterations to show in the figure. As can be seen from the figure, the loss value fluctuates slightly in the training process, but the overall trend is gradually decreasing. The loss value of view synthesis network decreased from the original 1.57 to the final 0.35, while the loss value of stereo matching network decreased from the initial 1.13 to the final 0.26.

\subsection{Evaluation Metrics}
\label{sect:Evaluation Metrics}
 We set the following parameters as the estimation metrics of the model and they demonstrate the error and performance of our method on depth evaluation using the ground truth depth data. The estimation metrics are used by Eigen et al.\cite{Eigen2014Depth} .

\begin{equation}\label{7}
    \begin{split}
         &RMSE(linear)=\sqrt{\frac{1}{N}\sum^N_{i=1}||Z_i-Z_i^{gt}||^2}\\
         &RMSE(log)=\sqrt{\frac{1}{N}\sum^N_{i=1}||log(Z_i)-log(Z_i^{gt})||^2}\\
         &Accuracy=\%Z_i:max(\frac{Z_i}{Z_i^{gt}},\frac{Z_i^{gt}}{Z_i})=\delta<thr\\
         &Abs\ Relative \ difference(ARD)=\frac{1}{N}\sum^N_{i=1}\frac{|Z_i-Z_i^{gt}|}{Z_i^{gt}}\\
         &Squared\ Relative\ difference(SRD)=\frac{1}{N}\sum^N_{i=1}\frac{||Z_i-Z_i^{gt}||^2}{Z_i^{gt}}\\
     \end{split}
\end{equation}

 Where N is the number of pixels about the ground truth depth map $Z^{gt}$ and evaluation depth map $Z$.

 To compare our method with current state-of-the-art methods of unsupervised monocular depth estimates and partial supervised monocular depth estimates,we crop our image resolution to match these models. Because these methods cap the evaluated depth in different ranges that Eigen et.al.\cite{Eigen2014Depth} and Godard et al.\cite{Godard2017Unsupervised} is 0-80m and Garg et.al \cite{Garg2016Unsupervised} is 1-50m, we respectively provide comparative results of the both depth distance. If the estimated depth value is outside the depth range, we set the depth value to be the lowest or highest value of the depth range.

\subsection{Results}

\textbf{Comparison of view synthesis network model:}
 The quality of the synthesized right view is very important for the depth estimation accuracy of the stereo matching network, so we set the evaluation formula for the output of the synthesized view network. We compute mean absolute error(MAE) between synthesized right view and original right view.
\begin{equation}\label{8}
  Mean \ Absolute \ Error(MAE)=\frac{1}{N}\sum_{i=1}^N|I_i^r-\bar{I}_i^r|
\end{equation}

 Where N is the number of pixels that was product of the image width and height. $I_i^r$ is the original right view and $\bar{I}_i^r$ is the synthesized right view. To evaluate our accuracy of view synthesis network model, we compare to a variant of this method based on the original unsupervised Deep3D \cite{Xie2016Deep3D} model and an improved one with adding smoothness constraint or modified $L1$ loss. Table ~\ref{tab:comp_Syn} shows the comparative results of different model.

\begin{table}[ht]\footnotesize
  \caption{Comparison of different view synthesis network models. For the Deep3D model,we uses the original model and an improved deep3d model with an added smoothness constraint(SC) and loss of L1+SSMI. The last row is the evaluation index of our model that outperform other model.}
  \label{tab:comp_Syn}
\setlength{\tabcolsep}{0.3mm}{
  \begin{center}
    \begin{tabular}{|c|c|c|c|c|c|c|c|c|c|}
    \hline
    \multirow{2}*{Method}&\multirow{2}*{Dataset}&MAE&RMSE&RMSE(log)&ARD&SRD&$\delta<1.25$&$\delta<1.25^2$&$\delta<1.25^3$\\
    \cline{3-10}
    &&\multicolumn{5}{|c|}{--lower is better--}&\multicolumn{3}{|c|}{--higher is better--}\\
    \hline
    \cline{1-1} \cline{3-10}
    Deep3D\cite{Xie2016Deep3D}&\multirow{4}*{K}&6.87&13.693&0.512&0.412&16.37&0.690&0.833&0.891\\
    \cline{1-1} \cline{3-10}
    Deep3D\cite{Xie2016Deep3D} with SC and L1+SSMI&\multirow{4}*{}&3.12&6.211&0.220&0.123&1.321&0.841&0.936&0.973\\
    \cline{1-1} \cline{3-10}
    Ours&\multirow{4}*{}&\textbf{3.02}&\textbf{6.096}&\textbf{0.214}&\textbf{0.120}&\textbf{1.300}&\textbf{0.846}&\textbf{0.939}&\textbf{0.980}\\
    \hline
    \end{tabular}
  \end{center}}

\end{table}

\begin{figure}
\begin{center}
    \begin{tabular}{c}
        \includegraphics[height=5.5cm]{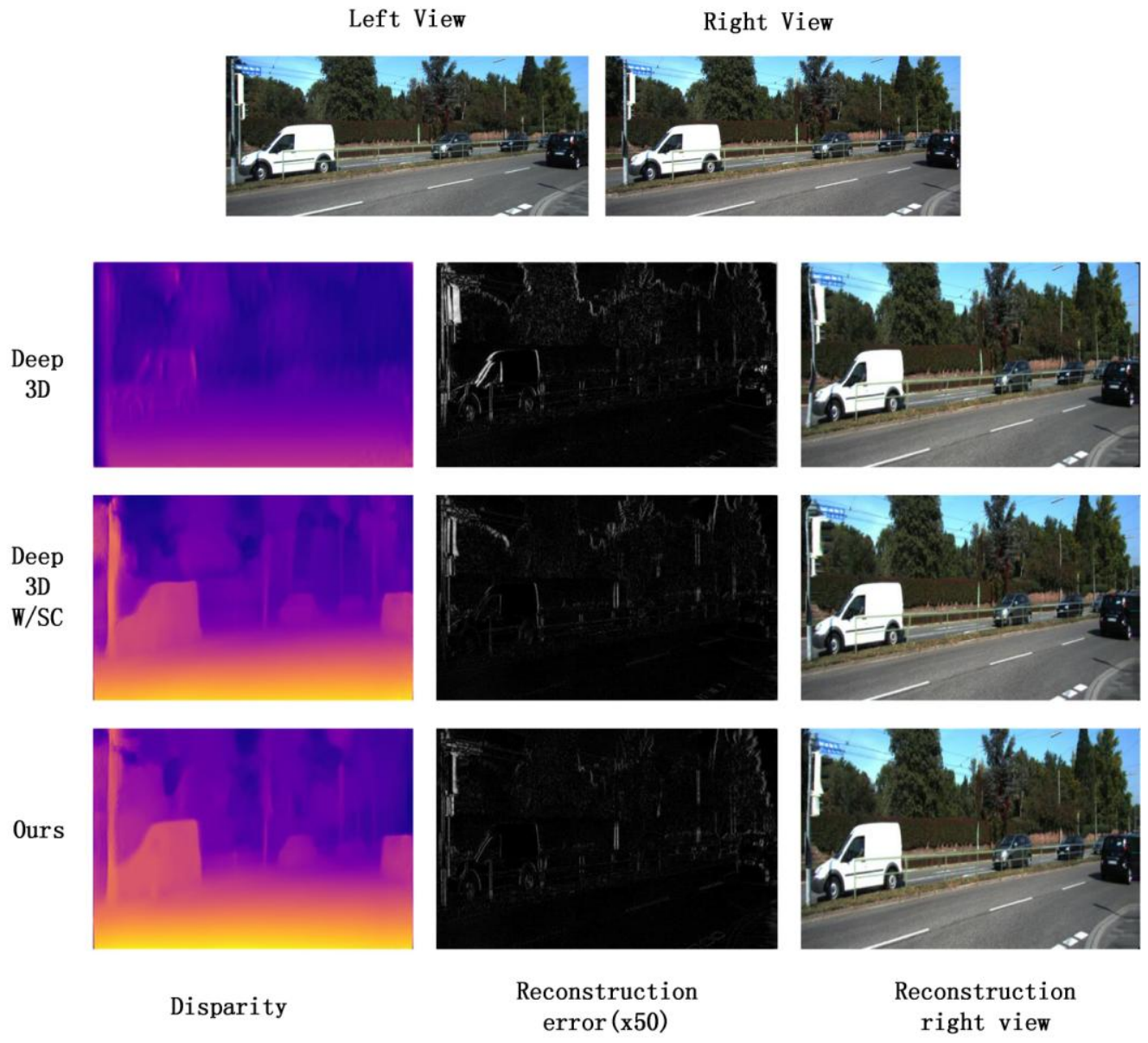}\\
    \end{tabular}
\end{center}
  \caption{The disparity, reconstruction error and reconstruction right view of the three models outputting, where the deep3D w/sc was the deep3d model with adding smoothing constraints and L1+SSMI loss functions}\label{fig:rec_loss}
\end{figure}

 As can be seen from Table ~\ref{tab:comp_Syn}, for deep3D model, the image reconstruction accuracy is greatly improved by adding smoothing constraints and L1+SSMI loss functions. Our model with additional Left-Right Disparity Consistency Loss outperform improved deep3D model 0.1 levels in MAE metrics for reconstructed right view. As the reconstruction process of right view is the result of the joint action of original left view and disparity value, disparity value plays an important role in view reconstruction. In order to compare our model with deep3D series model more accurately, the comparative results of model outputting disparity value according to sec.~\ref{sect:Evaluation Metrics} evaluation metrics is shown in table 2. As seen from the table, our model is better than Deep3D series model. Fig.~\ref{fig:rec_loss } visually shows the disparity, reconstruction error and reconstruction right view of the three models outputting.

\textbf{Comparisons with depth estimation:}
 Table 4 shows the comparison results of estimated depth value between our model and the current state-of-the-art unsupervised monocular depth evaluation method on the test dataset of the KITTI benchmark. As can be seen from the Table ~\ref{tab:Metreval_depth}, compared with other models, our model method has better results. In order to adapt to depth caps of various models, we conducted experiments on depth caps of 80m and 50m respectively for our models and compared them with other corresponding models. For the evaluation depth cap of 80m, our evaluation metrics outperform the unsupervised model of Godard et al.\cite{Godard2017Unsupervised} . In particular, the root mean square error(RMSE) and the square relative deviation(SRD) are better than the model 0.081m and 0.0774 respectively. However, our method is slightly inferior to this model in accuracy by 0.001m. When we compared our model with other method at an evaluation depth caps of 50m, we can see from Table ~\ref{tab:Metreval_depth} that our model, in depth evaluation metrics, are superior to the results reported by Godard et al.\cite{Godard2017Unsupervised} in almost every indicators and we can get the same performance as this model on the accuracy of the first power. However, compared with the results reported by Grag et al.\cite{Garg2016Unsupervised} model, we lost 0.242 in the root mean square error(RMSE), but we won by a large margin in other evaluation metrics. In general,the estimation depth caps of 50m is better than that of 80m in terms of the evaluation metrics.

\begin{figure}
\begin{center}
    \begin{tabular}{c}
        \includegraphics[height=3.5cm]{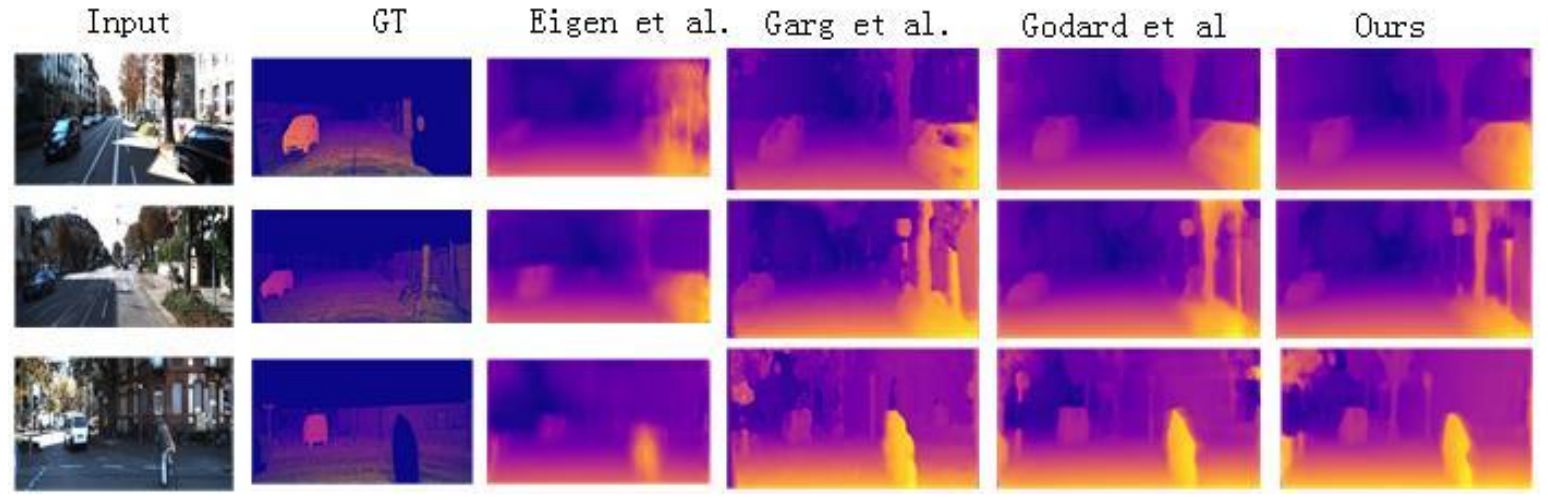}\\
    \end{tabular}
\end{center}
  \caption{Qualitative results about different methods on test dataset.This figure shows the depth estimation result from partial supervised model, current state-of-the-art unsupervised model and ground truth depth by the form of a visual image.}
  \label{fig:comp_depth}
\end{figure}

 We also compare our unsupervised model with the state-of-the-art unsupervised methods in terms of the output visual quality as Fig.~\ref{fig:comp_depth}. As we can see from the figure, the model of Godard et al.\cite{Godard2017Unsupervised} has been able to extract the depth map of car, person and traffic sign in the scene, but our method can present the details of the depth map more clearly and smoothly. There is no doubt that the ground truth depth data obtained by radar can present a better depth map.

\begin{table}[ht]\footnotesize
  \caption{Metrics evaluation results of our model and the current mainstream depth estimation model on the test of KITTI dataset using the split of Eigen et al.\cite{Eigen2014Depth}. This table shows two different caps of 50m and 80m between ground truth and estimated depth. where the SSD means that both the input data and the supervised data contain the synthesized right view, the OOD means that original right view as these two kinds of data, and that the SOD was that input data contain synthesized right view and the supervised data include original right view. However, in the test stage, mixed stereo pair is used as the input of the training models. We bold out the best results.}
  \label{tab:Metreval_depth}
  \begin{center}
  \setlength{\tabcolsep}{0.2mm}{
  \centering
    \begin{tabular}{|c|c|c|c|c|c|c|c|c|c|c|}
    \hline
    \multirow{2}*{Method}&\multirow{2}*{Dataset}&\multirow{2}*{Supervised}&\multirow{2}*{Cap}&RMSE&RMSE(log)&ARD&SRD&$\delta<1.25$&$\delta<1.25^2$&$\delta<1.25^3$\\
    \cline{5-11}
    &&&&\multicolumn{4}{|c|}{--lower is better--}&\multicolumn{3}{|c|}{--higher is better--}\\
    \hline
    \cline{1-1}
    Eigen et al. \cite{Eigen2014Depth} Coarse&\multirow{4}*{K}&Y&0-80m&6.215&0.271&0.204&1.598&0.695&0.897&0.960\\
    \cline{1-1} \cline{3-11}
    Eigen et al. \cite{Eigen2014Depth} Fine& &Y&0-80m&6.138&0.265&0.195&1.531&0.734&0.904&0.966\\
    \cline{1-1} \cline{3-11}
    Godard et al. \cite{Godard2017Unsupervised} & &N&0-80m&5.764&0.203&0.114&1.246&\textbf{0.854}&0.947&0.979\\
    \cline{1-1} \cline{3-11}
    Ours (SSD) & &N&0-80m&5.725&0.203&0.113&1.240&0.850&0.946&0.979\\
    \cline{1-1} \cline{3-11}
    Ours (OOD) & &N&0-80m&5.710&0.202&0.1127&\textbf{1.166}&0.850&0.946&\textbf{0.980}\\
    \cline{1-1} \cline{3-11}
    Ours (SOD) & &N&0-80m&\textbf{5.683}&\textbf{0.201}&\textbf{0.111}&1.1686&\textbf{0.854}&\textbf{0.948}&\textbf{0.980}\\
    \Xhline{0.75pt}\cline{1-2} \cline{3-11}
    Garg et al.\cite{Garg2016Unsupervised}&\multirow{4}*{K}&N&0-50m&5.104&0.273&0.169&1.080&0.740&0.904&0.962\\
    \cline{1-1} \cline{3-11}
    Godard et al. \cite{Godard2017Unsupervised} & &N&0-50m&5.431&0.199&0.110&1.034&\textbf{0.854}&0.949&0.980\\
    \cline{1-1} \cline{3-11}
    Ours (SSD)& &N&0-50m&5.413&0.199&0.109&0.975&0.850&0.949&0.980\\
    \cline{1-1} \cline{3-11}
    Ours (OOD)& &N&0-50m&5.404&0.198&0.110&0.973&0.850&0.948&\textbf{0.981}\\
    \cline{1-1} \cline{3-11}
    Ours (SOD)& &N&0-50m&\textbf{5.346}&\textbf{0.196}&\textbf{0.108}&\textbf{0.963}&\textbf{0.854}&\textbf{0.950}&\textbf{0.981}\\
    \hline
    \end{tabular}}
\end{center}
\end{table}

\begin{figure}
\begin{center}
    \begin{tabular}{c}
        \includegraphics[height=3.5cm]{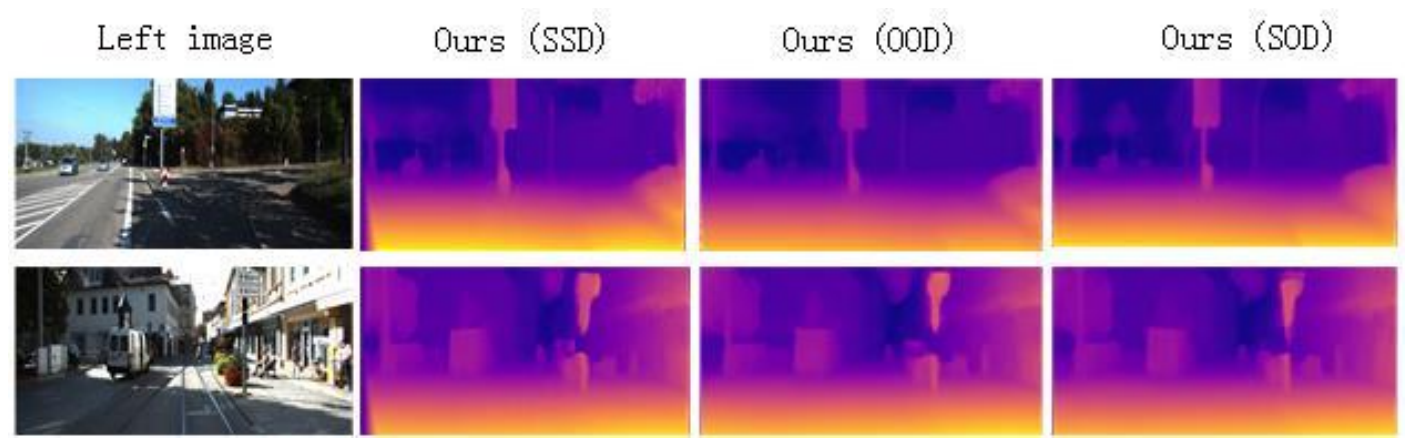}\\
    \end{tabular}
\end{center}
  \caption{The visual comparative result from different input data of stereo matching network at the training stage. where the SSD means that both the input data and the supervised data contain the synthesized right view, the OOD means that original right view as these two kinds of data, and that the SOD was that input data contain synthesized right view and the supervised data include original right view. }
  \label{fig:diff_stereo}
\end{figure}

\textbf{The influence of input data on stereo matching network in training stage:}
 As there is an error between the synthetic right view and the original right view, it can be seen from Table ~\ref{tab:comp_mon_bin} and Table ~\ref{tab:Metreval_depth} that the network model trained based on the synthetic right view is inferior to the network model trained based on the original in terms of estimation metrics. In order to verify the influence of training data on stereo matching network, we use three different input data to train the network model,which are respectively represented as the SSD, the OOD and the SOD in Table ~\ref{tab:Metreval_depth}. Where, in the training phase, the SSD refers to the original left view and synthesized right view as the stereo pairs, the OOD means that we use the original stereo pairs as as the stereo pairs, and the SOD refers to the original left view and synthesized right view as the stereo pairs. However, in the testing phase, mixed stereo pairs is used as the input of the three training models. As can be seen from table 3, no matter the evaluation caps is 50m or 80m, the training mode of the SOD is better than the other two training modes, and the evaluation metrics of the SSD training mode is the worst, even lower than monocular depth estimation methods of Godard et al.\cite{Godard2017Unsupervised} . At the depth cap of 80m, the performance of the OOD and the SOD are not different, even exceeding 0.003m in terms of the SRD metrics. When the depth cap is 50m, although the SOD is not much better than the OOD, it is clearly better than the OOD.

 Fig. ~\ref{fig:diff_stereo} shows the visual output results of the stereo matching models feeding in three input data. We can see from figure that since the synthetic right view is used as the supervision label for the training method of SSD, the depth map generated a large error in the edge part of the image. Although the OOD's result is better than that of SSD, the method does not refer to the synthetic right view during training, so the depth map generated is not optimal. For the method of SOD, We adapt mixed stereo pairs including synthetic right view as the input data and original stereo pairs as the supervision label,that can better optimize our network model.

\section{Conclusions}

In this paper, we propose an unsupervised stereo matching model based on monocular vision. The experimental results show that the stereo matching models outperforms monocular depth estimation models under the same unsupervised depth estimation model. Therefore, we first proposed a deep neural network to synthesize the right view from a single left view, and then used the reconstructed right view and original left view as the input of the unsupervised stereo matching network to estimate the depth. Both the view synthesis network and the stereo matching network are the unsupervised methods by only supervising on the image alignment loss from synthesized view and original view without the ground truth depth. Experimental results show that our method is superior to the current mainstream method of unsupervised depth estimation.

Although our method has certain advantages for some unsupervised methods, it cannot surpass the state-of-the-art supervised methods with the ground truth depth and the unsupervised methods with the original stereo pairs as the stereo matching method input. So you can see that our stereo matching network relies on a high-quality input data, we need to make further improvements to the view synthesis network to reconstruct the higher-quality right view. Finally, we may need a more appropriate network model to address the inconsistent depth of the problem caused by specular and transparent surfaces.


\bibliography{First_Paper}   

\begin{thebibliography}{10}

\bibitem{Furukawa2015Multi}
Y.~Furukawa, {\em Multi-View Stereo: A Tutorial}, Now Publishers Inc.  (2015).

\bibitem{L'2015Learning}
L.~Ladick{\'{y}}, H.~Christian, and M.~Pollefeys, ``Learning the matching
  function,'' {\em Computer Science}   (2015).

\bibitem{Sturm1996A}
P.~Sturm and B.~Triggs, ``A factorization based algorithm for multi-image
  projective structure and motion,'' {\em Proceedings Eccv96 London} {\bf
  1065}(3), 709--720  (1996).

\bibitem{Chen2016Single}
W.~Chen, Z.~Fu, D.~Yang, {\em et~al.}, ``Single-image depth perception in the
  wild,''  (2016).

\bibitem{Eigen2015Predicting}
D.~Eigen and R.~Fergus, ``Predicting depth, surface normals and semantic labels
  with a common multi-scale convolutional architecture,'' in {\em IEEE
  International Conference on Computer Vision},  2650--2658  (2015).

\bibitem{Ashutosh2009Make3D}
A.~Saxena, M.~Sun, and A.~Y. Ng, ``Make3d: Learning 3d scene structure from a
  single still image,'' {\em IEEE Trans Pattern Anal Mach Intell} {\bf 31}(5),
  824--840  (2009).

\bibitem{Lecun2016Stereo}
Y.~Lecun, {\em Stereo matching by training a convolutional neural network to
  compare image patches}, JMLR.org  (2016).

\bibitem{Luo2016Efficient}
W.~Luo, A.~G. Schwing, and R.~Urtasun, ``Efficient deep learning for stereo
  matching,'' in {\em Computer Vision and Pattern Recognition},  5695--5703
  (2016).

\bibitem{Ladick2014Pulling}
L.~Ladicky, J.~Shi, and M.~Pollefeys, ``Pulling things out of perspective,'' in
  {\em IEEE Conference on Computer Vision and Pattern Recognition},  89--96
  (2014).

\bibitem{Li2015Depth}
B.~Li, C.~Shen, Y.~Dai, {\em et~al.}, ``Depth and surface normal estimation
  from monocular images using regression on deep features and hierarchical
  crfs,'' in {\em IEEE Conference on Computer Vision and Pattern Recognition},
  1119--1127  (2015).

\bibitem{Laina2016Deeper}
I.~Laina, C.~Rupprecht, V.~Belagiannis, {\em et~al.}, ``Deeper depth prediction
  with fully convolutional residual networks,'' 239--248  (2016).

\bibitem{Garg2016Unsupervised}
R.~Garg, K.~B.~G. Vijay, G.~Carneiro, {\em et~al.}, ``Unsupervised cnn for
  single view depth estimation: Geometry to the rescue,'' 740--756  (2016).

\bibitem{Godard2017Unsupervised}
C.~Godard, O.~M. Aodha, and G.~J. Brostow, ``Unsupervised monocular depth
  estimation with left-right consistency,'' in {\em Computer Vision and Pattern
  Recognition},  6602--6611  (2017).

\bibitem{Zhou2017Unsupervised}
T.~Zhou, M.~Brown, N.~Snavely, {\em et~al.}, ``Unsupervised learning of depth
  and ego-motion from video,'' 6612--6619  (2017).

\bibitem{Luo2018Single}
Y.~Luo, J.~Ren, M.~Lin, {\em et~al.}, ``Single view stereo matching,''  (2018).

\bibitem{Ashutosh20083}
A.~Saxena, S.~H. Chung, and A.~Y. Ng, ``3-d depth reconstruction from a single
  still image,'' {\em International Journal of Computer Vision} {\bf 76}(1),
  53--69  (2008).

\bibitem{Pang2017Cascade}
J.~Pang, W.~Sun, J.~S. Ren, {\em et~al.}, ``Cascade residual learning: A
  two-stage convolutional neural network for stereo matching,'' 878--886
  (2017).

\bibitem{Zagoruyko2015Learning}
S.~Zagoruyko and N.~Komodakis, ``Learning to compare image patches via
  convolutional neural networks,'' in {\em IEEE Conference on Computer Vision
  and Pattern Recognition},  4353--4361  (2015).

\bibitem{Mayer2016A}
N.~Mayer, E.~Ilg, P.~H\"{a}usser, {\em et~al.}, ``A large dataset to train
  convolutional networks for disparity, optical flow, and scene flow
  estimation,'' in {\em Computer Vision and Pattern Recognition},  4040--4048
  (2016).

\bibitem{Long2015Fully}
J.~Long, E.~Shelhamer, and T.~Darrell, ``Fully convolutional networks for
  semantic segmentation,'' in {\em IEEE Conference on Computer Vision and
  Pattern Recognition},  3431--3440  (2015).

\bibitem{Dosovitskiy2015FlowNet}
A.~Dosovitskiy, P.~Fischery, E.~Ilg, {\em et~al.}, ``Flownet: Learning optical
  flow with convolutional networks,'' in {\em IEEE International Conference on
  Computer Vision},  2758--2766  (2015).

\bibitem{Eigen2014Depth}
D.~Eigen, C.~Puhrsch, and R.~Fergus, ``Depth map prediction from a single image
  using a multi-scale deep network,'' in {\em International Conference on
  Neural Information Processing Systems},  2366--2374  (2014).

\bibitem{Kaiming2015Deep}
K.~He, X.~Zhang, S.~Ren, {\em et~al.}, ``Deep residual learning for image
  recognition,'' 770--778  (2015).

\bibitem{Liu2016Learning}
F.~Liu, C.~Shen, G.~Lin, {\em et~al.}, ``Learning depth from single monocular
  images using deep convolutional neural fields,'' {\em IEEE Transactions on
  Pattern Analysis {\&} Machine Intelligence} {\bf 38}(10), 2024--2039  (2016).

\bibitem{Xie2016Deep3D}
J.~Xie, R.~Girshick, and A.~Farhadi, ``Deep3d: Fully automatic 2d-to-3d video
  conversion with deep convolutional neural networks,''  (2016).

\bibitem{Kuznietsov2017Semi}
Y.~Kuznietsov, J.~St¨¹ckler, and B.~Leibe, ``Semi-supervised deep learning for
  monocular depth map prediction,'' 2215--2223  (2017).

\bibitem{simonyan2014very}
K.~Simonyan and A.~Zisserman, ``Very deep convolutional networks for
  large-scale image recognition,'' {\em arXiv preprint arXiv:1409.1556}
  (2014).

\bibitem{Zhao2015Is}
H.~Zhao, O.~Gallo, I.~Frosio, {\em et~al.}, ``Is l2 a good loss function for
  neural networks for image processing?,'' {\em Computer Science}   (2015).

\bibitem{Zhou2004Image}
Z.~Wang, A.~Bovik, H.~Sheikh, {\em et~al.}, ``Image quality assessment: from
  error visibility to structural similarity,'' {\em IEEE Trans Image Process}
  {\bf 13}(4), 600--612  (2004).

\bibitem{Silberman:ECCV12}
P.~K. Nathan~Silberman, Derek~Hoiem and R.~Fergus, ``Indoor segmentation and
  support inference from rgbd images,'' in {\em ECCV},   (2012).

\bibitem{Geiger2012Are}
A.~Geiger, ``Are we ready for autonomous driving? the kitti vision benchmark
  suite,'' in {\em Computer Vision and Pattern Recognition},  3354--3361
  (2012).

\bibitem{Geiger2013Vision}
A.~Geiger, P.~Lenz, C.~Stiller, {\em et~al.}, ``Vision meets robotics: The
  kitti dataset,'' {\em International Journal of Robotics Research} {\bf
  32}(11), 1231--1237  (2013).

\bibitem{Abadi2016TensorFlow}
M.~Abadi, A.~Agarwal, P.~Barham, {\em et~al.}, ``Tensorflow: Large-scale
  machine learning on heterogeneous distributed systems,''  (2016).

\bibitem{Karen2014Very}
K.~Simonyan and Z.~Andrew, ``Very deep convolutional networks for large-scale
  image recognition,'' 1409--1556  (2014).

\end{thebibliography}
\bibliographystyle{spiejour}   

\end{document}